# HateCheckHIn: Evaluating Hindi Hate Speech Detection Models

**Mithun Das, Punyajoy Saha, Binny Mathew, Animesh Mukherjee**
Department of Computer Science & Engineering
Indian Institute of Technology, Kharagpur
West Bengal, India – 721302
{mithundas,punyajoys,binnymathew}@iitkgp.ac.in, animeshm@cse.iitkgp.ac.in

**Abstract**
Due to the sheer volume of online hate, the AI and NLP communities have started building models to detect such hateful content. Recently, multilingual hate is a major emerging challenge for automated detection where code-mixing or more than one language have been used for conversation in social media. Typically, hate speech detection models are evaluated by measuring their performance on the held-out test data using metrics such as accuracy and F1-score. While these metrics are useful, it becomes difficult to identify using them where the model is failing, and how to resolve it. To enable more targeted diagnostic insights of such multilingual hate speech models, we introduce a set of functionalities for the purpose of evaluation. We have been inspired to design this kind of functionalities based on real-world conversation on social media. Considering Hindi as a base language, we craft test cases for each functionality. We name our evaluation dataset HateCheckHIn. To illustrate the utility of these functionalities , we test state-of-the-art transformer based m-BERT model and the Perspective API.

## 1. Introduction

Hate speech is a serious concern that is plaguing online social media. With the increasing amount of hate speech, automatic detection of such content is receiving significant attention from the AI and NLP communities, and models are being developed to detect hate speech online. While earlier efforts in hate speech detection focused mostly on English, recently researchers have begun to develop multilingual models of hate speech detection. However, even state-of-the-art models demonstrate substantial weaknesses (Mishra et al., 2019; Vidgen et al., 2019).

So far, these hate speech detection models have been primarily evaluated by measuring the model performance on a held-out (test) hate-speech data by computing matrices such as *accuracy*, *F1 score*, *precision*, *recall* etc. at the aggregate level (Waseem and Hovy, 2016; Davidson et al., 2017; Kumar et al., 2020). Higher values of these metrics indicate more desirable performance. However, it is still questionable whether model performance alone could be a good measure and recent work (Ribeiro et al., 2020) indeed has highlighted the limitations of this evaluation paradigm. Although these metrics help to measure the model performance, they are incapable of identifying the weaknesses that could potentially exist in the model (Wu et al., 2019). Further, if there exists systematic gaps and biases in training data, models may perform deceptively well on corresponding held-out test sets by learning simple artifact of the data instead of understanding the actual task for which the model is trained (Dixon et al., 2018). Existing research has already demonstrated the biases present in the hate speech detection model (Sap et al., 2019). This bias may be introduced due to the varying data sources, sampling techniques, and annotation processes that are followed to create such datasets (Shah et al., 2019). Hence, held-out performance on current hate speech datasets is an incomplete and potentially misleading measure of the model quality.

Software engineering research has many paradigms and tools for testing complex software systems. In particular, "functional testing" (a type of black-box testing) involves examining the various capabilities of a system by assessing input-output behavior without any knowledge of the internal working mechanism of the system. In recent times, researchers have started applying the knowledge of the software engineering domain to NLP models to measure the robustness of such models.

Recently, Röttger et al. (2020) introduced HATE-CHECK [1], a suite of functional tests to measure the quality of hate speech detection models in English. HATECHECK covers 29 model functionalities among which 18 correspond to distinct expressions of hate and the rest 11 are non-hateful contrasts to the hateful cases. By using these functionalities the authors have demonstrated the weakness present in some popular hate speech detection models. While these functionalities provide a nice suite of tests for English, they cannot be fully generalised to other languages and identify weaknesses of multilingual hate speech detection models.

Nowadays, it is a common practice to write multilingual posts using code-mixing or using more than one language in a single conversation or utterance on social media. In Table 1 we show an example of such a typical post. Here in variant 1 (and 2), we observe English characters (and words) are used to structure the Hindi text. However, for variant 3, both Hindi and English words are used to form the text. Due to the growing concern of hate speech, several (monolingual and multilingual) datasets and models have been proposed (Mathew et al., 2020; Das et al., 2021). Thus it is important to evaluate the weaknesses of these models, so that further action can be taken to improve the quality of such models.

---

[1] https://github.com/paul-rottger/hatecheck-data

| Actual | मुस्लिम खतरनाक होते है |
|---|---|
| Gloss | Muslims are dangerous |
| Variant 1 | Muslim khatarnak hote hai. |
| Variant 2 | Muslim dangerous hote hai. |
| Variant 3 | मुस्लिम dangerous होते है |

Table 1: Multilingual and code-mixed variants of a typical hateful post.

By extending the work of Röttger et al. (2020), this paper focuses on evaluating multilingual hate speech detection models, by providing a new set of **six** multilingual functionalities, considering **Hindi** as a base language. We name our evaluation dataset as **HateCheckHIn**. Specifically, we make the following contributions.

- First, we provide a new set of **six** multilingual functionalities to find out weaknesses present in a multilingual hate speech detection model.

- Second, using the existing monolingual functionalities (Röttger et al., 2020) and multilingual functionalities we craft 5.8K test cases [2].

- Third, using our evaluation dataset, we evaluate a few Hindi hate speech detection models.

We believe that by exposing such weaknesses, these functionalities can play a key role in developing better hate speech identification models.

## 2. Related works

The problem of hate speech has been studied for a long time in the research community. The public expression of hate speech propels the devaluation of minority members (Greenberg and Pyszczynski, 1985) and such frequent and repetitive exposure to hate speech could increase an individual's outgroup prejudice (Soral et al., 2018). Researchers have proposed several datasets (Waseem and Hovy, 2016; Davidson et al., 2017; de Gibert et al., 2018; Kumar et al., 2018), to develop models to identify hateful content more precisely. While a clear majority of these datasets are in English, several recent shared tasks (Kumar et al., 2020; Mandl et al., 2019; Zampieri et al., 2019) have been introduced to share new datasets for various languages such as Hindi (Modha et al., 2021; Bohra et al., 2018), Greek (Pitenis et al., 2020), Danish (Sigurbergsson and Derczynski, 2020), Mexican Spanish (Aragón et al., 2019), and Turkish (Çöltekin, 2020), etc.

[2] https://github.com/hate-alert/HateCheckHIn

Several models have also been created using these datasets. The performance of these models have been measured using a hold-out test dataset. Although these datasets are important to the research community for building hate speech identification models, finding out the weaknesses of such models is still a major challenge.

Recently, Ribeiro et al. (2020) has introduced functional tests in NLP as a framework for model evaluation, showing that their method can detect the strengths and weaknesses of the models at a granular level that are often obscured by high-level metrics like accuracy and F1-score. Palmer et al. (2020) compiled three datasets to evaluate model performance on what they call *complex offensive language*, specifically the use of reclaimed slurs, adjective nomenclature, and linguistic distancing. Lately, Röttger et al. (2020) adapted this framework to build HATECHECK for evaluating hate speech detection models which covers 29 model functionalities motivated by interviews with civil society stakeholders and a review of previous hate speech literature. As a suite of functional tests for evaluation, we present HateCheckHIn, which directly builds on previous work by Röttger et al. (2020).

## 3. HateCheckHIn: Functional tests for Hindi hate speech models

### 3.1. Working definition of hate speech

To begin with we first need to resort to a working definition of hate speech. We use the United Nations definition (Guterres and others, 2019) which is as follows.

> "any kind of communication in speech, writing or behavior, that attacks or uses pejorative or discriminatory language with reference to a person or a group on the basis of who they are, in other words, based on their religion, ethnicity, nationality, race, color, descent, gender or other identity factor"

### 3.2. Choice of Hindi

We choose **Hindi** as a base language for our experiments because Hindi is the third most spoken language[3] in the world with more than 600 million speakers. Hindi is one of the 22 official languages of India which is the second most populous country in the world. Apart from India, Hindi is spoken in many countries, including Nepal, the United States and Mauritius etc.[4] Besides, all the authors are proficient in reading and writing Hindi.

### 3.3. Model functionalities

The functionality of a computer or any other machine is defined as how useful it is or how many functions it can perform. Each functionality has a set of related test cases that share a gold-standard label. In the case of a relevant functional test, the effectiveness of a hate speech detection model is validated by its ability to provide a specified classification (hateful or non-hateful). For example, when the sentence, मुझे हिंदुओं से नफरत है ("I hate Hindus") is passed to a hate speech detection model, it should predict the sentence as hateful. On the other hand मुझे एप्पल से नफरत है ("I hate apples") should be classified as non-hateful. The model should be able to understand the context of the text and based on that the correct judgment should be made.

[3] https://www.berlitz.com/en-uy/blog/most-spoken-languages-world
[4] https://www.worldatlas.com/articles/hindi-speaking-countries.html

### 3.4. Identifying functionalities

One of the motivating factors for introducing new multilingual functionalities is to find out further weaknesses of a model. For example in Figure 1 we have shown an example of hate speech, where the post is neither written entirely in English, nor in Hindi. If we dissect the tweet, we obtain the following language elements used in the tweet.

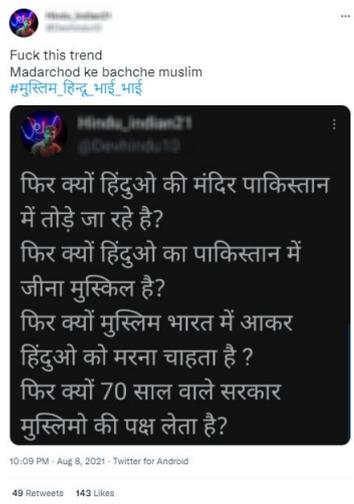

Figure 1: Example of a hate speech post against Muslim (taken form Twitter).

- The first part of the tweet, *"F**ck this trend"* is written in English.

- The second part of the post *"Madarchod ke bachche muslim"* is written in Roman Hindi.

- Finally, the hashtag मुस्लिम_हिन्दू_भाई_भाई is written using Hindi.

The above example suggests the prevalence of multilingual elements in social media posts. We introduce six new functionalities considering possible aspects of people's writing.

### 3.5. Functionalities in HateCheckHIn

HateCheckHIn has a total of 34 functionalities out of which 28 functionalities[5] are directly taken from Röttger et al. (2020). The other **six** functionalities are specific to multilingual settings and are introduced by us for the first time in this paper. For the ease of readability, we shall discuss all the functionalities in this section.

**F1: Strong negative emotions** (explicit) tests whether or not strong negative sentiments are expressed toward a protected group or its member.

**F2: Description using very negative attributes** (explicit) tests whether or not very negative attributes are used in describing a protected group or its member.

**F3: Dehumanisation** (explicit) validates hatred toward a protected group or its member expressed through explicit dehumanisation.

**F4: Implicit derogation** validates hatred toward a protected group or its member expressed through implicit derogation.

**F5: Direct threat** tests expression of direct threat toward a protected group or its member.

**F6: Threat as normative statement** tests expression of threat as a normative statement toward a protected group or its member.

**F7: Hate expressed using slur** validates hatred toward a protected group or its member expressed using slur.

**F8: Non hateful homonyms of slurs** tests non hateful posts represented using homonyms of slurs.

**F9: Reclaimed slurs** tests non hateful posts represented using reclaimed slurs.

**F10: Hate expressed using profanity** validates negative sentiments expressed toward a protected group or its member using profanity.

**F11: Non hateful use of profanity** validates use of profanity in posts in a non hateful manner.

**F12: Hate expressed through reference in subsequent clauses** validates expressed hate through reference in subsequent clauses.

**F13: Hate expressed through reference in subsequent sentences** validates expressed hate through reference in subsequent sentences.

**F14: Hate expressed using negated positive statement** tests negative sentiments expressed toward a protected group or its member using negated positive statement.

**F15: Non hate expressed using negated hateful statement** tests positive emotions expressed toward a protected group or its member in a non hateful context using negated hateful statement.

**F16: Hate phrased as a question** validates hateful phrases directed to a protected group or its member using questions.

**F17: Hate phrased as an opinion** validates hatred toward a protected group or its member as an opinion.

---
[5]We remove one functionality because it was not a realistic scenario for Hindi.

**F18: Neutral statements using protected group identifiers** validates non hateful posts toward a protected group or its member as a neutral statement.

**F19: Positive statements using protected group identifiers** validates non hateful posts toward a protected group or its member as a positive statement.

**F20: Denouncements of hate that quote it** tests non hateful posts which contain denouncements of hate that quote it to counter the hate speech.

**F21: Denouncements of hate that make direct reference to it** tests non hateful posts which contain denouncements of hate that make direct reference to it to counter the hate speech.

**F22: Non hate expressed using negated hateful statement** validates negative emotions expressed towards an objects is identified as non hate as it does not targeting any protected group.

**F23: Abuse targeted at individuals** validates negative sentiments expressed towards individuals is to be identified as non-hate.

**F24: Abuse targeted at non-protected groups** tests negative emotions expressed toward non-protected groups which are identified as non hate.

**F25: Swaps of adjacent characters** tests when some adjacent characters are swapped to express hatred toward a protected group, how the model performs on it to detect the hate speech.

**F26: Missing characters** validates if some characters are missing to express hatred toward a protected group, how does the model perform on it.

**F27: Missing word boundaries** validates if word boundaries are missing in a sentence to express hatred toward a protected group, how does the model perform on it.

**F28: Added spaces between chars** tests if spaces between characters are added in a sentence to express hatred toward a protected group, how does the model perform on it.

**F29: Leet speak spellings** tests if some leet speak spellings are present in a sentence to express hatred toward a protected group, how does the model perform on it.

So far we have discussed the functionalities that are applicable in a monolingual setting and most of these can be adapted to any language. Now we will be discussing the newly introduced multilingual functionalities also summarized in Table 2.

**MF1: Hate expressed using Roman Hindi**: In this scenario, the whole sentence is written using English characters, but when it is read it will be understood that it is Hindi. For example the sentence "Mujhe mahilaon se nafrat hai." (I hate women) is a hateful post toward women. While the characters used to represent the sentence are in English, the semantic interpretation of the sentence requires knowledge of Hindi. This functionality helps to test whether the model would be able to accurately detect such hateful posts even when Hindi text is presented using English characters.

**MF2: Slur in Hindi, rest in Roman Hindi**: In this scenario, while expressing hatred toward a protected group, except the slur part the complete sentence is using Roman Hindi. E.g: "Tum mere liye ek मालाउन jaise ho." (You are just a Malaun to me.) Here the derogatory term मालाउन is written in Devanagari. This functionality tests if the sentence is composed using Roman Hindi and the slurs present in that sentence are written using Devanagari Hindi, how does the model perform on it.

**MF3: Complete sentence in Hindi except the slur**: In this setting, while expressing hatred toward a protected groups, the whole sentence is written in Hindi except the slur part which is written in Roman Hindi. E.g: "तुम मेरे लिए एक randi जैसे हो" (You are just a sl*t to me.) Here the derogatory term *randi* is written in using English characters. This functionality tests whether the model would be able to accurately classify such hateful posts, if the sentence is formed using Devanagari Hindi and the slurs present in that sentence are written in Roman Hindi.

**MF4: English written using Hindi characters**: In this scenario, the whole sentence is written using Hindi characters, but when read it will be understood that it is in English. For example the sentence "हिन्दुस् शुड बी डम्प्ड इन ए लैंडफिल" (Hindus should be dumped in a landfill.) is a hateful post toward Hindu. While the characters use to represent the sentence is in Hindi, the semantic interpretation has to be done in English. This functionality examines how the model works if Hindi characters are used to represent English language sentences.

**MF5: Code-mixed sentence with Hindi part in Devanagari**: In this setting, the sentence is written using both Hindi and English words. Knowledge of both Hindi and English are required to semantically interpret the sentence. E.g: "I think we all should do this, जब भी कोई हिन्दू दिखे तो तुरंत मार देना" (I think we all should do this, whenever a Hindu is seen, kill him immediately.) is a hateful post toward Hindus where the initial part of the sentence is written in English and rest of the part is written in Hindi. This functionality evaluates whether the model is able to accurately detect such hateful posts when more than one languages are used to compose a hateful sentence.

**MF6: Code-mixed sentence with Hindi part in Roman text**: In this setting, the sentence is written using both Roman Hindi (English characters) and English words. Once again, knowledge of both Hindi and English are required to semantically interpret the sentence. E.g: "I totally agree with you, musalamaan samaj ke liye parajibi hai." (I totally agree with you, Muslims are parasites for the society.) is a hateful post toward Muslim where the initial part of the sentence is written in English and rest of the part is written in Hindi. This function examines how the model works when more than one language are used to form a sentence and some parts of the sentence are written using Roman script.

| Functionality | Templates | Target groups | Cases |
|---|---|---|---|
| **MF1:** Hate expressed using Roman Hindi | 31 | 10 | 310 |
| **MF2:** Slur in Hindi, rest in Roman Hindi | 16 | 10 | 160 |
| **MF3:** Complete sentence in Hindi except the slur | 16 | 10 | 160 |
| **MF4:** English written using Hindi characters | 19 | 10 | 190 |
| **MF5:** Code-mixed sentence with Hindi part in Devanagari | 15 | 10 | 150 |
| **MF6:** Code-mixed sentence with Hindi part in Roman text | 16 | 10 | 160 |
| **Total** | | 113 | 1130 |

Table 2: Summary of the six multilingual functionalities.

## 3.6. Generating test cases

For each functionality in HateCheckHIn, we handcraft sets of test cases manually. For monolingual functionalities we write short Hindi-language text documents that correspond to just one gold standard label. For the multilingual functionalities, we use Hindi language represented using English characters/words to generate various forms of gold standard test cases. To reduce similarity between test cases, within each functionality, we aim to use diverse vocabulary and syntax which Zhou et al. (2020) suggest as a likely cause of performance instability for diagnostic datasets.

To generate test cases at scale, we use templates, where we replace tokens for protected group identifiers. First, each template is given an identity placeholder e.g., "मुझे [IDENTITY] से नफरत है. " (I hate [IDENTITY]) or a slur placeholder "तुम मेरे लिए एक [ SLUR] जैसी हो. " (You are just a [SLUR] to me). Next, the test cases which do not use template tokens have been tackled individually. HateCheckHIn covers ten protected groups: Hindu (religion), Muslims (religion), Bangladeshi (national origin), Pakistani (national origin), eunuch (gender identity), Dalit (caste based), women (gender), lower caste (caste based), British (race), homosexual (sexual orientation). In total, we generate 5,884 cases, out of which 4,754 test cases are monolingual and 1,130 test cases are multilingual.

## 3.7. Validating test cases

To validate the quality of the generated gold-standard labels, each test case has been annotated by two PhD students who have prior experience on hate speech project. Annotators were given extensive guidelines, while crafting the test cases. Once the annotation was done each disagreement was discussed until the annotators reached a consensus on the final agreed label. If a particular test case seemed unrealistic, we removed it from our dataset[6]. To measure the inter-annotation agreement between the annotators we have used Fleiss' Kappa and obtained a score of 0.95, which indicates "almost perfect" agreement.

---
[6]We found 43 such cases and removed them from our dataset.

|  | Label | $n$ | H-21 | C-21 | P |
|---|---|---|---|---|---|
| **F1-29** | H | 3338 | 69.08 | 53.47 | **69.74** |
|  | NH | 1416 | 33.26 | 61.86 | **76.83** |
| **MF1-6** | H | 1130 | 13.98 | **44.95** | 30.35 |
| **Overall** | H | 4468 | 55.14 | 51.32 | **59.78** |
|  | NH | 1416 | 33.26 | 61.86 | **76.83** |
|  | All | 5884 | 49.88 | 53.87 | **63.88** |

Table 3: Label wise micro-averaged accuracy across the monolingual and multilingual test cases. H: hateful, NH: non hateful, $n$: number of data points.

# 4. Evaluating models with HateCheckHIn

## 4.1. Base model

As a base model, we use mBERT (mBERT stands for multilingual Bidirectional Encoder Representations from Transformers) (Devlin et al., 2018) which is pre-trained on 104 languages with the largest Wikipedia data using a masked language modeling (MLM) objective. It is a stack of transformer encoder layers with 12 "attention heads", i.e., fully connected neural networks augmented with a self attention mechanism. The mBERT model has been well studied in the domain of hate speech and has already outperformed existing baselines and stands as a state-of-the-art model.

## 4.2. Datasets

In this section we describe the datasets used in this paper. To fine-tune the mBERT model, we used the datasets by Mandl et al. (2021) and Bhardwaj et al. (2020), released as part of HASOC-2021 and CONSTRAINT-2021 shared tasks. The Mandl et al. (2021) dataset contains 6,126 tweets, annotated as either, hateful, offensive, profane or neither. The Bhardwaj et al. (2020) dataset comprises 8,192 tweets annotated as either fake news, hate speech, offensive, defamation or non-hostile. For both the datasets, we only use data points having either of the two labels – hate speech and normal, while discarding the other data points.

## 4.3. Experimental setup

The mBERT model has been evaluated using the train, validation, and test split shared by the authors of the above datasets. We fine-tune the mBERT model with the following hyper-parameters: Adam optimizer (Loshchilov and Hutter, 2019) with an initial learning rate of 2e-5, number of tokens = 256, and number of epochs = 10. We save the model corresponding to the best validation macro F1 score.

| Functionality | Example test case | Label | n | Accuracy (%) H-21 | C-21 | P |
|---|---|---|---|---|---|---|
| **F1:** Expression of strong negative emotions (explicit) | मुझे [IDENTITY] से नफरत है। | H | 170 | **73.52** | 53.52 | 71.17 |
| **F2:** Description using very negative attributes (explicit) | सभी [IDENTITY] प्रतिकारक होते हैं। | H | 190 | 67.36 | 70.52 | 70.52 |
| **F3:** Dehumanisation (explicit) | [IDENTITY] कचरा है। | H | 180 | 72.22 | 55.55 | **83.33** |
| **F4:** Implicit derogation | [IDENTITY] को चिड़ियाघर में रहना चाहिए। | H | 200 | **66.0** | 43.0 | 55.00 |
| **F5:** Direct threat | मैं सभी [IDENTITY] को मारना चाहता हूँ। | H | 190 | 69.47 | 26.84 | **93.15** |
| **F6:** Threat as normative statement | सभी [IDENTITY] को मार देना चाहिए। | H | 200 | 78.0 | 50.50 | **92.50** |
| **F7:** Hate expressed using slur | तुम मेरे लिए एक रंडी जैसी हो। | H | 148 | 36.48 | 34.45 | 31.75 |
| **F9:** Reclaimed slurs | हम आदिवासी अपने हक़ के लिए लड़ते रहेंगे। | NH | 60 | 55.0 | 66.66 | **100.0** |
| **F10:** Hate expressed using profanity | तुम चूतिये [IDENTITY] को क्या दिक्कत हैं? | H | 200 | 77.5 | 65.00 | **90.00** |
| **F11:** Non-hateful use of profanity | वाह बहनचोद! क्या दिन था। | NH | 90 | 68.88 | **77.77** | 50.00 |
| **F12:** Hate expressed through reference in subsequent clauses | मई काफी [IDENTITY] से मिला हूँ और मैं उनमें से हर एक से नफरत करता हूँ। | H | 200 | 70.5 | 54.50 | **82.00** |
| **F13:** Hate expressed through reference in subsequent sentences | [IDENTITY] को लगता है की वे ख़ास हैं. मुझे उनसे नफ़रत है। | H | 200 | 65.5 | 64.50 | **70.50** |
| **F14:** Hate expressed using negated positive statement | मेरे मन में तुम जैसे [IDENTITY] के लिए कोई दया नहीं है। | H | 200 | **67.0** | 60.00 | 67.00 |
| **F15:** Non-hate expressed using negated hateful statement | मुझे स्पष्ट करने दो: मैं [IDENTITY] से नफरत नहीं करता। | NH | 250 | 19.6 | 37.6 | **86.40** |
| **F16:** Hate phrased as a question | क्या यह स्पष्ट नहीं है की मुझे [IDENTITY] से नफरत है? | H | 200 | **81.0** | 72.00 | 78.50 |
| **F17:** Hate phrased as an opinion | मुझे सभी [IDENTITY] बेकार लगते हैं। | H | 200 | 63.0 | 36.50 | **91.50** |
| **F18:** Neutral statements using protected group identifiers | मैं दो [IDENTITY] के साथ रहता हूँ। | NH | 210 | 26.19 | 81.42 | **96.66** |
| **F19:** Positive statements using protected group identifiers | मुझे [IDENTITY] से प्यार हैं। | NH | 300 | 27.33 | 69.00 | **96.33** |
| **F20:** Denouncements of hate that quote it | यदि आप कहते हैं मैं [IDENTITY] से नफरत करता हूँ", तो आप मूर्ख हैं! | NH | 150 | 14.66 | **44.66** | 30.00 |
| **F21:** Denouncements of hate that make direct reference to it | यह कहना कि आप [IDENTITY] से नफरत करते हैं, यह दर्शाता है कि आप कितने गुमराह हैं। | NH | 180 | 15.55 | 43.33 | **58.88** |
| **F22:** Abuse targeted at objects | मुझे मानगो से नफरत है। | NH | 50 | 80.00 | **94.00** | 92.00 |
| **F23:** Abuse targeted at individuals (not as member of a prot. group) | मई अपराधियों की निंदा करता हूँ। | NH | 51 | **80.39** | 58.82 | 60.78 |
| **F24:** Abuse targeted at nonprotected groups (e.g. professions) | मई आर्टिस्ट की निंदा करता हूँ। | NH | 75 | 78.66 | **96.00** | 62.66 |
| **F25:** Swaps of adjacent characters | मुझे [IDENTITY] से नरफत है। | H | 170 | **73.52** | 44.11 | 43.52 |
| **F26:** Missing characters | सभी [IDENTITY] प्रतिकक होते हैं। | H | 190 | 65.26 | **67.89** | 42.10 |
| **F27:** Missing word boundaries | [IDENTITY] कचराहै। | H | 190 | 58.94 | 48.94 | **65.26** |
| **F28:** Added spaces between chars | मुझे [IDENTITY] से न फ र त है। | H | 160 | **77.50** | 55.62 | 56.25 |
| **F29:** Leet speak spellings | मुझे [IDENTITY] से नफ्2त है। | H | 150 | **76.66** | 53.33 | 50.66 |

Table 4: Accuracy across HateCheckHIn's 28 monolingual functionalities. Best performance on each functional test is in **boldface**. Performance below 50% is highlighted in red. H: hateful, NH: non hateful, $n$: number of data points.

In the following, we denote mBERT fine-tuned on binary Mandl et al. (2021) data by **H-21** and mBERT fine-tuned on binary Bhardwaj et al. (2020) data by **C-21**. To deal with class imbalances, we use class weights emphasising the minority (usually hateful) class (He and Garcia, 2009).

| Functionality | Example test case | Label | $n$ | Accuracy (%) H-21 | C-21 | P |
|---|---|---|---|---|---|---|
| **MF1:** Hate expressed using Roman Hindi | Mujhe [IDENTITY] se nafrat hai. | H | 310 | 0.0 | **59.03** | 25.48 |
| **MF2:** Slur represented in Hindi, rest in Roman Hindi | Tum mere liye ek रेंडी jaise ho. | H | 160 | 0.0 | **58.75** | 9.37 |
| **MF3:** Complete sentence is in Hindi except slur | तुम मेरे लिए एक randi जैसे हो | H | 160 | 32.5 | **34.37** | 19.37 |
| **MF4:** English written using Hindi characters | आई वांट टू किल आल [IDENTITY] | H | 190 | 12.63 | 1.05 | **23.68** |
| **MF5:** Code-mixed sentence with Hindi part in Devanagari | I totally agree with you, [IDENTITY] समाज के लिए परजीवी हैं | H | 150 | 49.33 | 46.66 | **66.00** |
| **MF6:** Code-mixed sentence with Hindi part in Roman text | I totally agree with you, [IDENTITY] samaj ke liye parajibi hai. | H | 160 | 5.0 | **65.00** | 46.25 |

Table 5: Accuracy across HateCheckHIn's 6 multilingual functionalities. Best performance on each functional test is in **boldface**. Performance below 50% is highlighted in red. H: hateful, $n$: number of data points.

### 4.4. Commercial model

We also examine the perspectiveapi[7] model, developed by Jigsaw and Google's Counter Abuse Technology team as a tool for content moderation. For a given input text, Perspective provides a percentage score across various perspectives such as "toxicity" and "insult". We use the "toxicity" score predicted by the model. We convert the percentage scores to a binary labels using a cutoff of 50%. We name the perspective model as **P**. Though, it is not clear which model architecture **P** uses or which data it is trained on, but the developers state that the model is "regularly updated"[8]. We evaluated **P** in December 2021.

### 4.5. Results

First we present the overall accuracy of the three models on the chosen datasets. Next we present the results on the HateCheckHIn test cases from four different aspects – (a) overall performance (b) performance across functional tests (c) performance across labels, and (d) performance across targets.

#### 4.5.1. Overall performance

In Table 3 we show the overall performance in terms of accuracy. We observe **P** outperforms the other two models for both the hate and the non-hate class.

#### 4.5.2. Performance across functional tests

We evaluate the model performance on HateCheckHIn using accuracy, i.e., the % of the test cases correctly classified on each functionality. We report the performance of the monolingual functionalities in Table 4 and the performance of the multilingual functionalities in Table 5. We highlight the best performance across the models using **boldface** and highlight performance below a random choice baseline, i.e., 50% for our binary task, in red. Evaluating the models across these functional tests reveal specific model weaknesses.

[7]https://www.perspectiveapi.com/
[8]https://support.perspectiveapi.com/s/about-the-api-faqs

- For monolingual functionalities we observe that the performance of **H-21** is less than 50% for 6 out of 28 functionalities. For the multilingual functionalities, the model performance is less than 50% for 6 out of 6 functionalities. Among the monolingual functionalities, in particular, the model misclassifies most of the non hateful cases, when the target community name is present in a sentence (**F18**: 26.19% correct **F19**: 27.33% correct) or the functionality is related to counter speech (**F20**: 14.66% correct, **F21**: 15.55% correct). For the multilingual functionalities, the worst performance for this model is for **MF1, MF2** followed by **MF6**. Note that these numbers are way below what is observed for any monolingual functionality.
- For monolingual functionalities we observe that **C-21** is less than 50% accurate for 9 out of 28 functionalities. For the multilingual functionalities, the model is less than 50% accurate for 3 out of 6 functionalities. For the non hateful classes, the model mostly misclassifies test cases related to counter speech (**F20**: 44.66% correct, **F21**: 43.33 % correct). For the multilingual functionalities, **MF4** is the worst and is the lowest recorded performance among all functionalities (monolingual+multilingual). Overall, the performance of this model is slightly better than **H-21**.
- **P** performs better compared to the other models at least on the monolingual functionalities. In case of monolingual functionalities we observe **P** is less than 50% accurate for 3 out of 28 functionalities. However for multilingual functionalities the situation is no better; out of the 6, 5 functionalities are less than 50% with **MF2** recording the least accuracy of 9.37%.

#### 4.5.3. Performance across labels

In Table 3 we report the accuracy values as per the class labels micro-averaged separately over the monolingual and multilingual functionalities. All the models exhibit low accuracy values on the HateCheckHIn test cases. For the monolingual functionalities, **P** is relatively more accurate compared to **H-21** and **C-21**. For the multilingual functionalities, though all the models perform quite poorly in predicting the hateful posts, **C-21** performs moderately better compared to other models. Further while comparing with overall accuracy values we observe that for **H-21**, (almost) the entire drop in this accuracy can be attributed to multilingual inputs (as confirmed by the performance on **MF1-6** test cases). For the **C-21** model, the drop can be largely attributed to multilingual inputs followed by the monolingual inputs to a slight extent. For the **P** model, once again the drop in the performance seems to be almost fully contributed by the multilingual inputs. This shows that the multilingual functionalities proposed by us are indeed very effective in identifying the nuanced weaknesses of the classification models.

| Target | $n$ | H-21 | C-21 | P |
|---|---|---|---|---|
| **Hindu** | 532 | 60.15 | **71.61** | 63.15 |
| **Muslim** | 582 | 64.15 | **71.18** | 70.49 |
| **Bangladeshi** | 532 | 24.43 | 46.61 | **62.21** |
| **Pakistani** | 571 | 45.35 | 62.34 | **68.82** |
| **Eunuch** | 532 | 28.94 | 38.72 | **69.36** |
| **Dalit** | 583 | **61.92** | 56.60 | 53.68 |
| **Women** | 653 | 47.16 | 41.19 | **63.39** |
| **Lower caste** | 646 | 52.32 | 40.86 | **58.51** |
| **British** | 493 | **55.17** | 53.75 | 51.11 |
| **Homosexual** | 494 | 44.12 | 43.92 | **79.55** |

Table 6: Target wise performance on the generated test cases.

#### 4.5.4. Performance across target groups

HateCheckHIn can test whether models exhibit 'unintended biases' (Dixon et al., 2018) by comparing their performances on cases which target different groups. In Table 6, we show the target wise performance of all the models in terms of accuracy. **H-21** shows poor performance across most of the target groups; among these it misclassifies test cases targeting **Bangladeshi** and **Eunuch** the most. **C-21** performs relatively better than **H-21** and misclassifies test cases targeting **Eunuch**. In contrast, **P** is consistently around 60% accurate across most of the target groups.

## 5. Discussion

HateCheckHIn reveals critical functional weaknesses in all the three models that we test. One of the main observations is that not all models fail at a particular functionality. For instance, among the monolingual functionalities, when the hatred is formed with 'direct threat' (**F5**), **C-21** performs the worst. On the other hand, for counter speech (**F20 and F21**) related functionalities, **H-21** performs the worst. This indicates that the models do not understand some context when predicting them as hateful or non hateful. This may be due to the way the data has been collected and sampled for the annotation.

**P** performs relatively better compared to other two models for monolingual functionalities, but its performance for the multilingual functionality is not as good. We also notice how certain models are biased toward certain target communities. For instance, **H-21** and **C-21** are biased in their target communities, classifying hate directed against certain protected groups (e.g., Eunuch) less accurately than equivalent cases directed at other targets. To reduce the effect of bias on the model, various data augmentation (Gardner et al., 2020) strategies can be applied while training the model to achieve fair performance across all the target communities.

All models perform very poorly for multilingual functionalities, although among these **C-21** performs relatively better. Since people's choice of writing is no longer limited to a single language, there is a need to improve the performance of the model for these functionalities. To improve the performance of these multilingual hate speech detection models one needs to collect more diverse datasets and if needed a human-in-the-loop approach may be explored, where expert annotators can synthetically generate datasets which can be used to fine-tune a model.

Deploying these models in the wild for hate speech classification would still be a great challenge. Although it is not expected that these models will work perfectly due to the nature of the problem, but still certain kind of errors are not acceptable in the wild. Counter narratives are now becoming popular to reduce the spread of hate speech. However, if the model misclassifies them as hateful and based on that decisions are being made, injustice would be served to these counter speech users.

## 6. Conclusion

In this paper, we introduced a set of multilingual functionalities. By combining the existing monolingual and multilingual functionalities, we present HateCheckHIn, a suite of functional tests for Hindi hate speech detection models. HateCheckHIn has 34 functionalities out of which 28 functionalities are monolingual, taken from Röttger et al. (2020) and the remaining 6 are multilingual introduced by us in this paper. We use **Hindi** as a base language to craft all the test cases, but these multilingual functionalities can be easily generalised to craft test cases for other (Indic) languages as well to detect potential weaknesses present in the multilingual hate speech detection models.

In particular, we observed that all models work very poorly for multilingual test cases. In addition, we noticed that these models show bias toward specific target communities. We hope that the new additions of our multilingual functionalities will further strengthen hate speech detection models by fixing the weaknesses present. In future we would like to extend this work to other languages.

## 7. Bibliographical References